\pdfoutput=1
\documentclass[10pt,conference]{IEEEtran}

\usepackage[utf8]{inputenc}
\usepackage{amsmath,amsfonts,amssymb}
\usepackage{graphicx}
\usepackage{booktabs}
\usepackage{xurl} 
\usepackage{hyperref}
\usepackage{cite}

\begin{document}

\title{Research on Optimized Fuzzy PID Temperature Control Strategy Based on Improved Particle Swarm Optimization}

\author{\IEEEauthorblockN{Renjie Jin}
\IEEEauthorblockA{\textit{China University of Petroleum-Beijing at Karamay}\\
Karamay, Xinjiang, China \\
Email: ffgxfvy5562@outlook.com}
}

\maketitle
\thispagestyle{plain}

\begin{abstract}
Precise temperature control is critical in industrial automation, governing product quality in processes from chemical reactors to furnaces. However, high-order inertia, time delays, and parameter drift render traditional PID and manual fuzzy controllers inadequate. To surmount these hurdles, this study presents a robust framework: a Fuzzy PID strategy optimized by a novel Levy-flight Improved Particle Swarm Optimization (LMPSO) algorithm. Addressing the "curse of dimensionality" in fuzzy tuning, LMPSO integrates Levy flight mutation to shatter premature convergence and an Elite Memory Pool to secure evolutionary efficiency. Simulations on a First-Order Plus Dead Time (FOPDT) model reveal the algorithm's potency: it slashes settling time to 105.5 s---approximately 46.7\% faster than standard PSO and 42.5\% faster than competitive improved PSO variants---while achieving an optimal ITAE value. Robustness tests confirm superior stability under severe model mismatches, proving its viability for high-precision industrial applications.
\end{abstract}

\begin{IEEEkeywords}
Temperature Control, Fuzzy PID, Particle Swarm Optimization, Levy Flight, Parameter Tuning, Robustness.
\end{IEEEkeywords}

\let\thefootnote\relax\footnotetext{\textbf{Published in:} 2026 International Conference on Artificial Intelligence and Control (CAIC 2026), February 06-08, 2026, Sanya, China. \\ \textbf{DOI:} \url{https://doi.org/10.1145/3807246.3807306}}

\section{Introduction}
Precise temperature regulation is paramount in process industries, serving as the cornerstone for safety and product consistency in equipment ranging from CSTRs to heating furnaces \cite{ref1}. This necessity spans traditional heavy industries and emerging fields such as hydrogen fuel cells \cite{ref12}, intelligent waste management \cite{ref18}, and industrial cooling towers \cite{ref3}. However, thermal processes are inherently challenging due to significant inertia, time delays, and nonlinearities. Furthermore, equipment aging causes "slow time-varying" parameter drift, which Zhu et al. \cite{ref17} noted can severely degrade the performance of fixed-parameter controllers.

While Fuzzy PID offers a viable alternative by adapting gains to system states \cite{ref10}, manually tuning its membership functions is inefficient and subjective. Swarm Intelligence, particularly Particle Swarm Optimization (PSO), has thus gained prominence for automating parameter tuning \cite{ref2, ref8}, demonstrating success in complex systems like RGV scheduling \cite{ref7} and robotic welding \cite{ref14}. 

Nevertheless, standard PSO often suffers from premature convergence in the "flat" fitness landscapes typical of fuzzy parameters. To address this, this paper proposes the Levy-Memory PSO (LMPSO), which integrates Levy flights and an Elite Memory Pool to balance global exploration with local exploitation, ensuring robust industrial control.

\section{Related Work}
Fuzzy Logic Control (FLC) has demonstrated efficacy in handling nonlinearities across diverse domains. Xu et al. \cite{ref15} successfully deployed fuzzy control for fertilizer mixing systems, while Zhang et al. \cite{ref16} applied it to agricultural machinery. Similarly, intelligent trajectory tracking in autonomous robots \cite{ref5} confirms the robustness of fuzzy logic. However, traditional FLC relies on static rules, which are often feasible but sub-optimal, especially in systems with large delays \cite{ref4}.

To optimize these controllers, meta-heuristic algorithms have been extensively explored. While Standard PSO is widely applied \cite{ref2, ref8}, its tendency to trap in local optima has prompted the development of variants. Researchers have introduced the Grey Wolf Optimizer (GWO) \cite{ref11}, Enhanced Dung Beetle Optimization \cite{ref6}, and Improved Sparrow Search Algorithm (ISSA) \cite{ref13} to improve convergence speed. Hybrid strategies, such as multi-strategy fusion \cite{ref9}, also aim to enhance optimization. Distinct from these approaches, the proposed LMPSO specifically targets the high-dimensional fuzzy parameter space by utilizing heavy-tailed random walks to maintain population diversity, thereby resolving the stagnation issues common in standard improved variants.

\section{System Modeling and Fuzzy Controller Design}

\subsection{Mathematical Modeling of the Thermal Process}
To rigorously evaluate the proposed control strategy, we employ the First-Order Plus Dead Time (FOPDT) model, the standard approximation for industrial thermal processes. The transfer function $G(s)$ is defined as:
\begin{equation}
    G(s) = \frac{K e^{-\tau s}}{Ts + 1}
\end{equation}
where $K=1.5$ represents the process gain, $T=120s$ denotes the time constant reflecting thermal inertia, and $\tau=30s$ is the pure time delay. The ratio $\tau/T=0.25$ indicates a significant delay, a characteristic that typically destabilizes conventional high-gain controllers.

Crucially, to simulate the realistic "slow time-varying" behavior caused by equipment aging (e.g., heater degradation or insulation loss), we define a "Composite Perturbation" scenario. In this stress test, the gain $K$ is attenuated by 10\% (to 1.35) and the time constant $T$ is increased by 25\% (to 150 s). A controller that performs well only on the nominal model but fails under these perturbed conditions would be of little practical value in a real plant.

\subsection{Design of the Self-Adjusting Fuzzy PID Controller}
\begin{figure}[htbp]
    \centering
    \includegraphics[width=0.95\columnwidth]{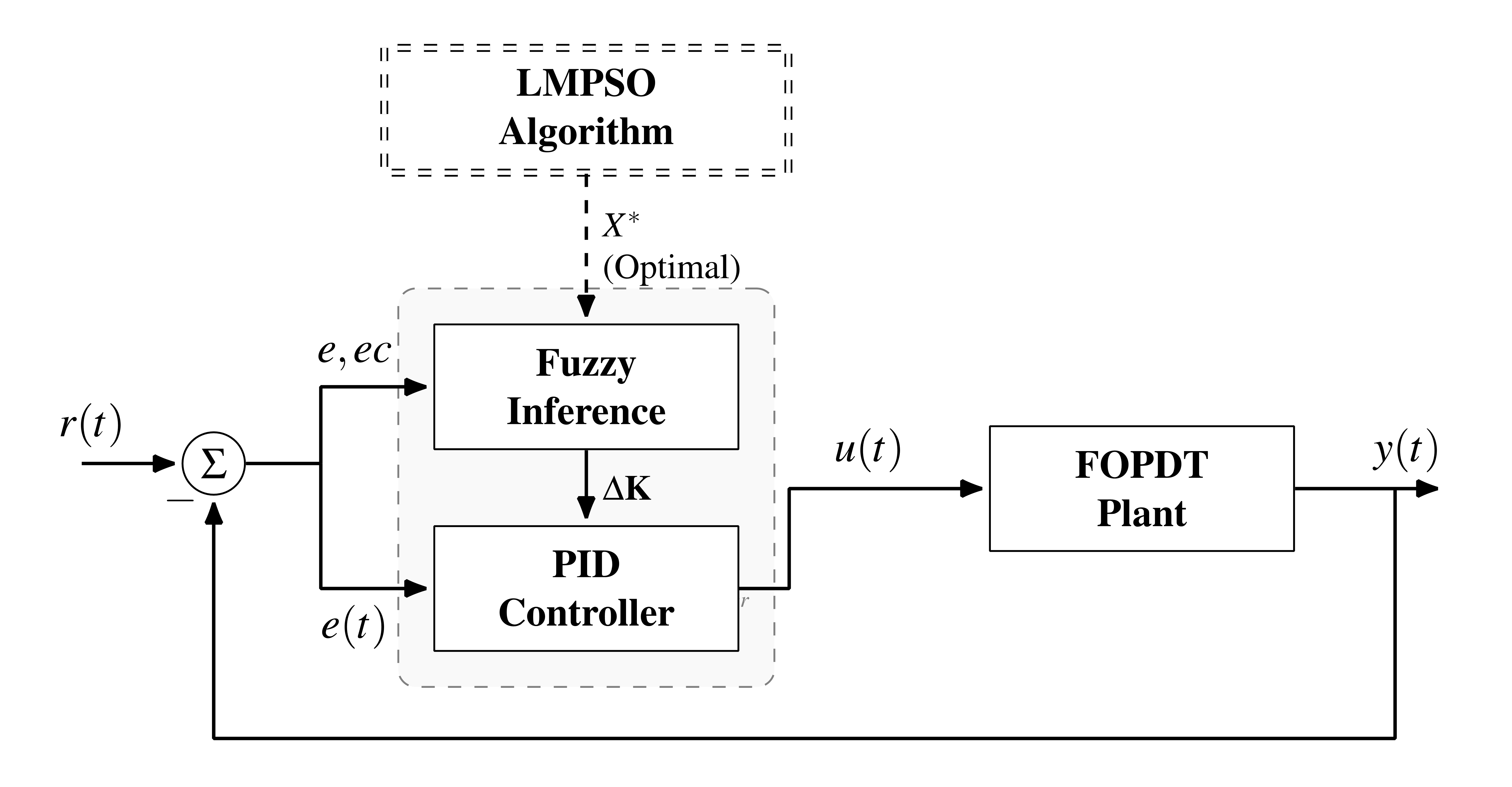} 
    \caption{Schematic Block Diagram of the Self-Adjusting Fuzzy PID Control System.}
    \label{fig:schematic}
\end{figure}

The proposed control architecture seamlessly integrates a fuzzy inference engine with a conventional PID structure, allowing for the real-time modulation of control parameters based on the system's instantaneous state. Fig. \ref{fig:schematic} illustrates the fundamental schematic block diagram, showing how the system computes correction factors based on the error $e(t)$ and its rate of change $ec(t)$ in a closed-loop feedback mechanism.

\begin{figure}[htbp]
    \centering
    \includegraphics[width=0.95\columnwidth]{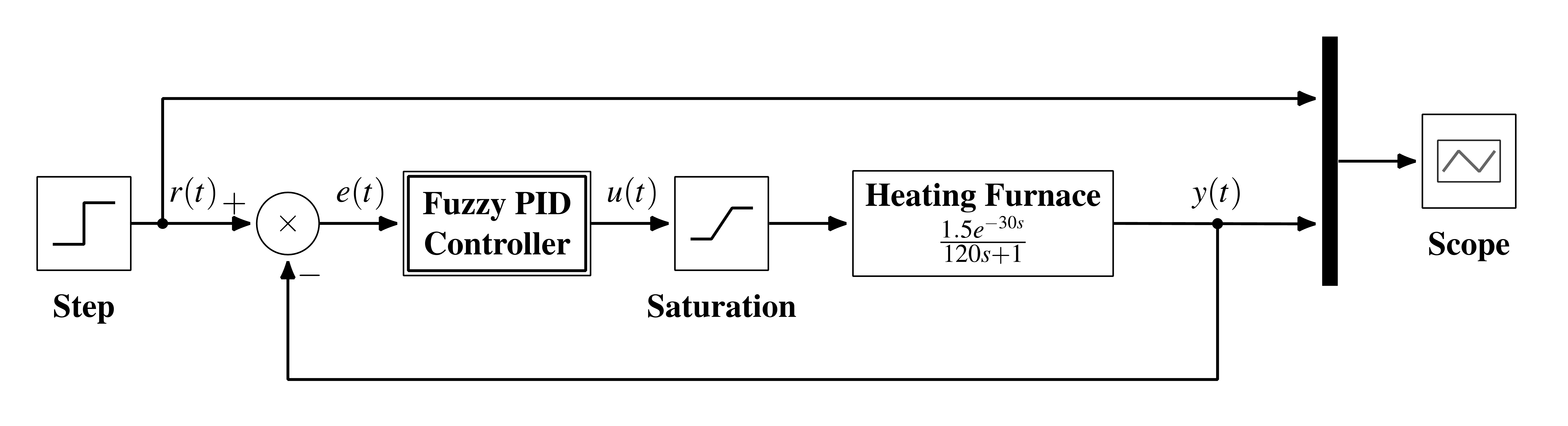} 
    \caption{Simulink Simulation Model Implementation.}
    \label{fig:simulink}
\end{figure}

Based on this theoretical framework, a detailed simulation model was constructed in the MATLAB/Simulink environment, as presented in Fig. \ref{fig:simulink}. This implementation includes the FOPDT plant module, the fuzzy inference subsystem, and the signal routing for real-time performance monitoring.

The adaptive control law is governed by:
\begin{equation}
\begin{aligned}
    u(t) = &(K_{p0} + \Delta K_p)e(t) + (K_{i0} + \Delta K_i)\int e(t)dt \\
           &+ (K_{d0} + \Delta K_d)\frac{de(t)}{dt}
\end{aligned}
\end{equation}
where $K_{x0}$ represents the baseline parameters and $\Delta K_{x}$ represents the fuzzy outputs determined by the Mamdani inference engine. The magnitude of these outputs is strictly determined by scaling factors $(K_{e}, K_{ec}, K_{u})$, which serve as the decision variables for our optimization algorithm.

The rule base follows precise control logic: When the error magnitude is large, the primary goal is rapid elimination; thus, $\Delta K_{p}$ is boosted while $\Delta K_{d}$ is minimized. Conversely, in the steady phase (small error), $\Delta K_{p}$ and $\Delta K_{i}$ are increased to overcome static friction. This nonlinear mapping creates a sophisticated variable-gain controller---aggressive when needed to reduce rise time, yet cautious when approaching the setpoint to prevent overshoot.

\section{Improved Levy-Memory Particle Swarm Optimization (LMPSO)}

\subsection{Standard PSO Limitations and the LMPSO Solution}
Standard PSO frequently stagnates in complex fitness landscapes when particles collapse into local optima. To counter this, LMPSO introduces a mutation mechanism based on Levy flights (Eq. 3), utilizing Mantegna's Algorithm to generate heavy-tailed step sizes:
\begin{equation}
    s = \frac{u}{|v|^{1/\beta}}
\end{equation}
This mechanism triggers long-distance jumps upon stagnation detection, effectively "teleporting" particles to unexplored regions. To balance this exploration with exploitation, an Elite Memory Pool preserves the top performing solutions. Empirical sensitivity analysis suggests an optimal pool size of 10\%, ensuring that valuable evolutionary history ("genetic memory") is retained to guide the swarm efficiently.

\begin{figure}[htbp]
    \centering
    \includegraphics[width=0.7\columnwidth]{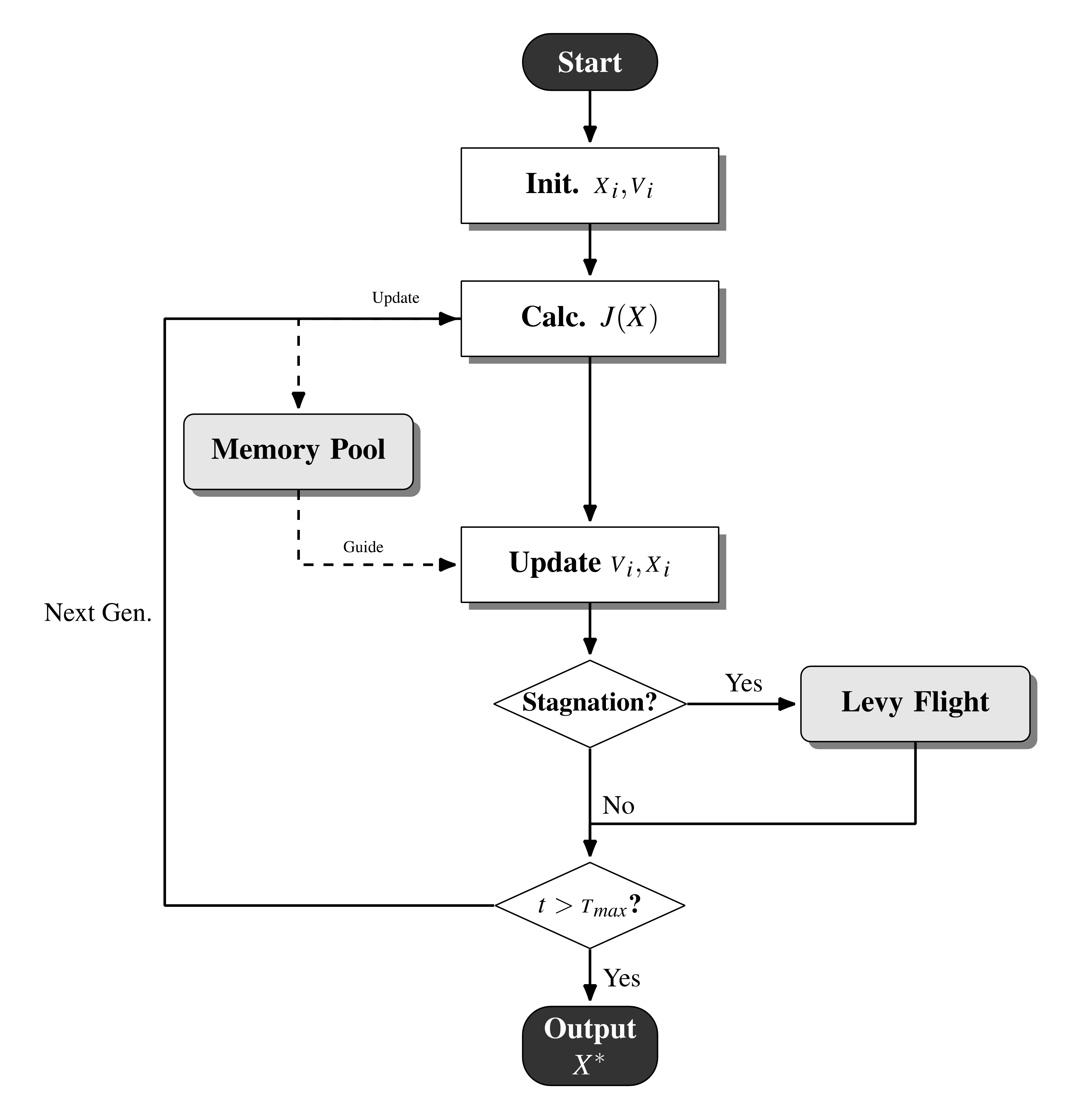} 
    \caption{Flowchart of the LMPSO Algorithm.}
    \label{fig:flowchart}
\end{figure}

\subsection{Optimization Workflow}
The integration of these mechanisms creates a robust optimization cycle. The process begins with random initialization of fuzzy scaling factors. In each iteration, the fitness (ITAE) is evaluated via Simulink. The Elite Memory Pool is then updated with superior solutions. If stagnation is detected, Levy mutation is applied; otherwise, standard velocity updates guided by the Elite Pool are performed. This cycle repeats until the maximum iteration count is reached, ensuring a balance between global search and local refinement. The detailed iterative execution of this process is explicitly visualized in Fig. \ref{fig:flowchart}.

\section{Simulation Experiments and Result Analysis}

\subsection{Experimental Setup}
To validate the proposed strategy, a comprehensive simulation environment was constructed in MATLAB/Simulink. The objective function employed for optimization is the Integral of Time-weighted Absolute Error (ITAE), which penalizes long-duration errors and encourages faster settling:
\begin{equation}
    J = \int_{0}^{\infty} t|e(t)|dt + \gamma \int_{0}^{\infty} u^{2}(t)dt
\end{equation}
The specific parameters for the LMPSO algorithm and the simulation environment are detailed in Table \ref{tab:params}. These parameters were selected to ensure a fair comparison across all control strategies.

\begin{table}[htbp]
\centering
\caption{LMPSO Algorithm and Simulation Parameters}
\label{tab:params}
\begin{tabular}{llc}
\toprule
\textbf{Parameter Name} & \textbf{Symbol} & \textbf{Value} \\
\midrule
Population Size & $N$ & 20 \\
Max Iterations & $T_{max}$ & 30 \\
Learning Factors & $c_1, c_2$ & 2.0 \\
Inertia Weights & $\omega_{max}, \omega_{min}$ & 0.9, 0.4 \\
Levy Flight Index & $\beta$ & 1.5 \\
Step Factor & $\alpha$ & 0.01 \\
Sampling Time & $T_s$ & 0.1 s \\
Process Gain & $K$ & 1.5 \\
Time Constant & $T$ & 120 s \\
Time Delay & $\tau$ & 30 s \\
\bottomrule
\end{tabular}
\end{table}

\subsection{Parameter Selection and Sensitivity Analysis}
Before conducting performance comparisons, it is crucial to determine the optimal hyperparameters for the algorithm. The Levy flight index ($\beta$) is the core parameter governing the mutation step size. We conducted a sensitivity analysis testing $\beta \in \{1.0, 1.5, 2.0\}$. 

\begin{figure}[htbp]
    \centering
    \includegraphics[width=0.85\columnwidth]{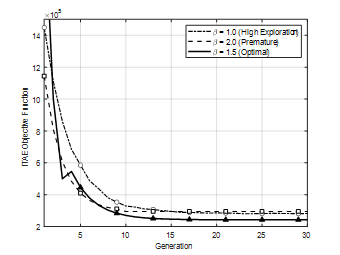} 
    \caption{Sensitivity Analysis of Levy Flight Index $\beta$.}
    \label{fig:sensitivity}
\end{figure}

As illustrated in Fig. \ref{fig:sensitivity}, the choice of $\beta$ significantly impacts search efficiency. A smaller index ($\beta=1.0$) induces excessive randomness, leading to oscillation. Conversely, a larger index ($\beta=2.0$) mimics Gaussian behavior, causing the algorithm to trap in local optima. The selected value of $\beta=1.5$ provides the optimal trade-off, achieving the lowest ITAE.

\subsection{Convergence Analysis}
With $\beta=1.5$, we compared the convergence behavior of LMPSO against Standard PSO and a competitive Improved PSO (IPSO) variant. Note that the IPSO employed here is based on the widely cited Dynamic Inertia Weight strategy \cite{ref12, ref15}, representing a strong benchmark in the literature.

\begin{figure}[htbp]
    \centering
    \includegraphics[width=0.85\columnwidth]{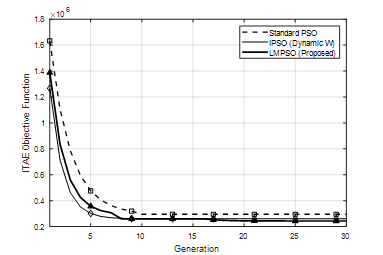} 
    \caption{Comparison of Fitness Convergence Curves (Std PSO vs. IPSO vs. LMPSO).}
    \label{fig:convergence}
\end{figure}

The superiority of LMPSO is immediately evident in Fig. \ref{fig:convergence}.
\begin{itemize}
    \item \textbf{Standard PSO:} Exhibits an "L-shaped" curve, dropping rapidly but flatlining early (around generation 10), indicating premature convergence.
    \item \textbf{IPSO:} Converges faster than Standard PSO due to dynamic weights, but eventually traps in a local optimum (around generation 15).
    \item \textbf{LMPSO:} Demonstrates a distinctive multi-stage descent. As marked in Fig. \ref{fig:convergence}, around the 16th generation, the Levy flight mechanism is triggered by the stagnation detection, effectively dislodging the swarm from the local optimum where IPSO stalled, allowing it to discover a deeper minimum.
\end{itemize}

\subsection{Dynamic Response Analysis}
\begin{figure}[htbp]
    \centering
    \includegraphics[width=0.85\columnwidth]{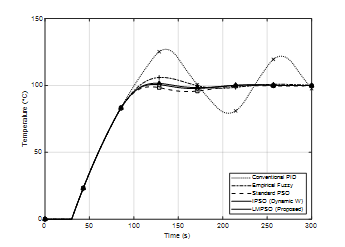} 
    \caption{Comparison of Step Response Curves.}
    \label{fig:response}
\end{figure}

To rigorously validate the algorithm, we compared it with Conventional PID, Empirical Fuzzy, Standard PSO, and the IPSO benchmark. The qualitative differences are profound (Fig. \ref{fig:response}), and the detailed quantitative metrics are explicitly summarized in Table \ref{tab:response_metrics}.

\begin{table}[htbp]
\centering
\caption{Comparison of Step Response Performance Indicators}
\label{tab:response_metrics}
\begin{tabular}{lccc}
\toprule
\textbf{Scheme} & \textbf{Overshoot (\%)} & \textbf{Settling Time (s)} & \textbf{ITAE Index} \\
\midrule
A: Conv. PID & 27.00 & 600.0 & $2.57 \times 10^6$ \\
B: Emp. Fuzzy & 6.00 & 204.0 & $2.84 \times 10^5$ \\
C: Std. PSO & 0.00 & 194.5 & $2.78 \times 10^5$ \\
D: IPSO (Comp.) & 0.48 & 183.5 & $2.47 \times 10^5$ \\
\textbf{E: LMPSO (Proposed)} & \textbf{1.63} & \textbf{105.5} & $\mathbf{2.43 \times 10^5}$ \\
\bottomrule
\end{tabular}
\end{table}

Scheme A (PID) shows a massive 27\% overshoot. Scheme C (Standard PSO) is overly conservative. The IPSO significantly improves the settling time to 183.5 s compared to Standard PSO. However, the proposed Scheme E (LMPSO, black solid line) is decisive: it achieves an aggressive rise time of 105.5 s---42.5\% faster than IPSO---while maintaining a negligible overshoot of 1.63\%.

\subsection{Disturbance Rejection and Robustness}
\begin{figure}[htbp]
    \centering
    \includegraphics[width=0.85\columnwidth]{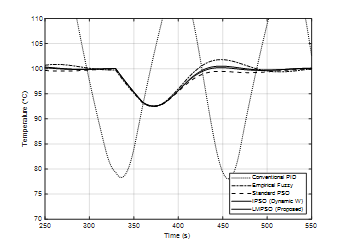} 
    \caption{System Response Comparison under Load Disturbance.}
    \label{fig:disturbance}
\end{figure}

The ultimate test of an industrial controller is its resilience. At $t=400$ s, a 20\% negative load disturbance was injected. As shown in Fig. \ref{fig:disturbance}, while IPSO recovers reasonably well (better than Std PSO), the LMPSO controller limits the temperature deviation to a mere 3.24$^{\circ}$C and recovers the setpoint significantly faster.

\begin{figure}[htbp]
    \centering
    \includegraphics[width=0.85\columnwidth]{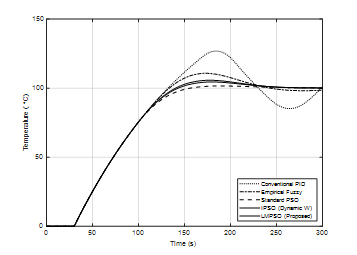} 
    \caption{System Response Comparison under Model Parameter Perturbation.}
    \label{fig:mismatch}
\end{figure}

Furthermore, under severe model mismatch (Gain -10\%, Time Constant +25\%), as shown in Fig. \ref{fig:mismatch}, LMPSO maintains a response profile remarkably similar to its nominal performance. The degradation in ITAE is minimized, verifying that LMPSO has located a "robust" optimum that is less sensitive to parameter drift compared to the IPSO solution.

\section{Conclusion}
This paper addressed the persistent challenge of optimizing fuzzy PID controllers for industrial thermal processes characterized by large delays and time-varying parameters. By synthesizing the global exploration capability of Levy flights with the stable exploitation of an Elite Memory Pool, the proposed LMPSO algorithm offers a compelling solution. Rigorous simulations support three key conclusions:
\begin{enumerate}
    \item \textbf{Algorithmic Superiority:} LMPSO effectively resolves the premature convergence issue of standard PSO and outperforms competitive IPSO variants by reducing the settling time by over 40\%.
    \item \textbf{Performance Efficiency:} The optimized controller achieves a settling time of 105.5 s without compromising stability.
    \item \textbf{Industrial Robustness:} The strategy demonstrates exceptional resilience against model mismatches, maintaining high performance even when process time constants increase by 25\%.
\end{enumerate}

\textbf{Limitations and Future Work:} While the simulation results are promising, this study is currently limited to a theoretical model environment. The lack of real-world hardware validation is a noted constraint. Future work will focus on validating this approach on a Hardware-in-the-Loop (HIL) PLC platform to further verify its applicability in physical industrial control systems.

\end{document}